\documentclass[lettersize,journal]{IEEEtran}
\usepackage{amsmath,amsfonts}
\usepackage{algorithmic}
\usepackage{array}
\usepackage[caption=false,font=normalsize,labelfont=sf,textfont=sf]{subfig}
\usepackage{textcomp}
\usepackage{stfloats}
\usepackage{url}
\usepackage{verbatim}
\usepackage{graphicx}
\def\BibTeX{{\rm B\kern-.05em{\sc i\kern-.025em b}\kern-.08em
    T\kern-.1667em\lower.7ex\hbox{E}\kern-.125emX}}
\usepackage{balance}

\usepackage{booktabs}       

\begin{document}
\title{P3S-Diffusion:A Selective Subject-driven Generation Framework via Point Supervision}
\author{
\IEEEauthorblockN{
		Junjie Hu \textsuperscript{1}, 
        Shuyong Gao \textsuperscript{1,}\IEEEauthorrefmark{1}, 
        Lingyi Hong \textsuperscript{1}, 
        Qishan Wang \textsuperscript{1}, \\
        Yuzhou Zhao \textsuperscript{1}, 
		Yan Wang \textsuperscript{1,2}
		and Wenqiang Zhang\textsuperscript{1,2,3,}\IEEEauthorrefmark{1}} 

	\IEEEauthorblockA{\textsuperscript{1}Shanghai Key Lab of Intelligent Information Processing,\\ School of Computer Science, Fudan University, Shanghai, China.\\ 
		}
	\IEEEauthorblockA{\textsuperscript{2}Shanghai Engineering Research Center of AI \& Robotics, Academy for Engineering \& Technology,\\ Fudan University, Shanghai, China.\\} 
    \IEEEauthorblockA{\textsuperscript{3}Engineering Research Center of AI \& Robotics, Ministry of Education, Academy for Engineering \& Technology,\\ Fudan University, Shanghai, China.\\ }
    
    \IEEEauthorblockA{
    \{jjhu23,lyhong22,yzzhao22\}@m.fudan.edu.cn \\
    \{sygao18,qswang20,yanwang19,wqzhang\}@fudan.edu.cn
    }
}


\maketitle

\def\OurFramework{P3S-Diffusion }
\def\OurEncoder{RNGR-Encoder }
\def\OurEncoderTitle{RNGR-Encoder}

\begin{figure*}
    \centering
    \includegraphics[width=1\linewidth]{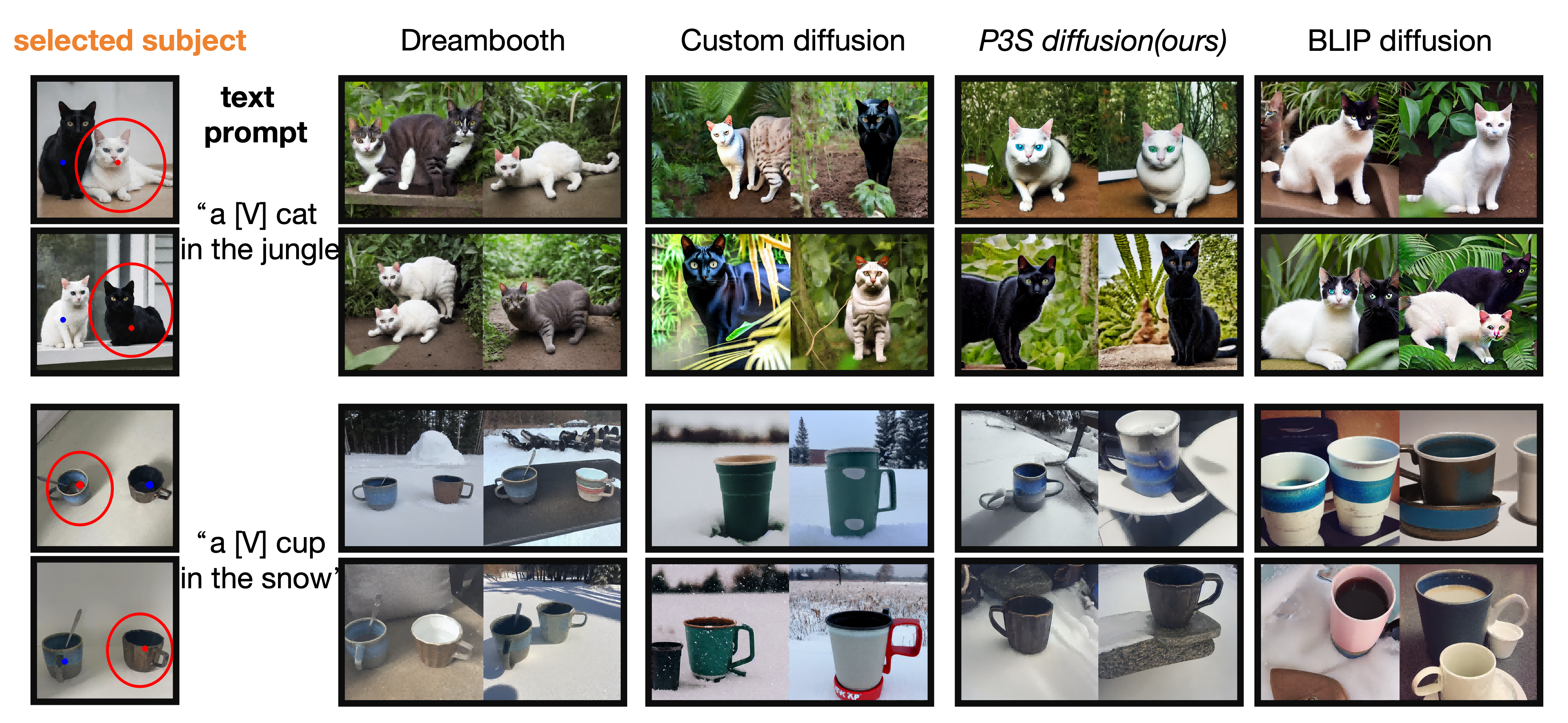}
    \caption{\textbf{Selective subject comparison}. Our results not only excel in selective subject generations but also performs well in terms of fidelity to the reference subjects.}
    \label{fig:point_comparison}
\end{figure*}

\begin{abstract}
Recent research in subject-driven generation increasingly emphasizes the importance of selective subject features. Nevertheless, accurately selecting the content in a given reference image still poses challenges, especially when selecting the similar subjects in an image (e.g., two different dogs). Some methods attempt to use text prompts or pixel masks to isolate specific elements. However, text prompts often fall short in precisely describing specific content, and pixel masks are often expensive. To address this, we introduce \OurFramework, a novel architecture designed for context-selected subject-driven generation via point supervision. \OurFramework leverages minimal cost label (e.g., points) to generate subject-driven images. During fine-tuning, it can generate an expanded base mask from these points, obviating the need for additional segmentation models. The mask is employed for inpainting and aligning with subject representation. The \OurFramework preserves fine features of the subjects through Multi-layers Condition Injection. Enhanced by the Attention Consistency Loss for improved training, extensive experiments demonstrate its excellent feature preservation and image generation capabilities.
\end{abstract}


\section{Introduction}
\IEEEPARstart{R}{ecently} Text-to-Image (T2I) generative models demonstrate the ability to generate high-quality and text-related images. However, relying solely on textual descriptions cannot accurately generate the images we want. Even if using large language models (LLM) to expand text descriptions to detailed ones, it is still not possible to generate images that meet the requirement. Furthermore, inversion techniques \cite{text-inversion} can help to find a text embedding but still can't accurately reconstruct the appearance of given subjects. So now there are many models \cite{control-net} dedicated to utilizing more control conditions (e.g., image prompt, sketch, depth map, etc.) to guide T2I diffusion models.
 
 Subject-driven generation aims to generate realistic and fidelity-preserving images where the subject is the same as the reference image. Given a few images of a particular subject (3-5) , models should accurately reconstruct the appearance of the given subjects using special prompts while retaining editability.

 In this work, we present a new approach named as \OurFramework, which generate \textbf{P}oint \textbf{S}upervision \textbf{S}elective \textbf{S}ubject-driven image for personalized generation using point labels to select target subject. Recent research \cite{control-net, ssrencoder} have tried to utilize extra conditions (e.g., additional text prompts and mask images) to accurately select the subject in the input image. However, text prompts often fail to describe specific content accurately, and pixel masks are always expensive. Therefore, we weaken the selection criteria and make it possible to use only points to guide diffusion model in learning special subjects. A novel Attention Consistency Loss is proposed to enhance the correlation between text prompts and generated images.

 Our \OurFramework consists of two parts:\OurEncoder and Multi-layers Condition Injection. We eliminate the confusion of similar subjects using \OurEncoder, a new image encoder that \textbf{R}educes \textbf{N}egative information and \textbf{G}enerates spatial bias \textbf{R}epresentations of the subject. \OurEncoder first utilizes CLIP \cite{CLIP} image encoder to get the patch-level relation of the subjects and reduce negative information. Furthermore, we propose a detailed subject condition injection, which utilizes a trainable copy of U-Net to inject multi-layers subject information. To strengthen editable generation capability, we propose the $\textit{timestep-based weight scheduler}$, which can flexibly adjust the weight and trend of control condition injection to balance  prompt consistency and identity preservation.
 We summarize our main contributions as follows:
 \begin{itemize}
 \item We propose a novel framework, named as \OurFramework for selective subject-driven image generation.Comparing to existing text prompts or mask supervision methods, our method requires the least additional control and can accurately select the target subject.It 
 can be flexibly applied to diffusion models and compatible with ControlNets.
 \item \OurEncoder and Multi-layers Condition Injection are proposed to select target subject and encode selected feature to original U-Net.We minimize the Attention Consistency Loss between original U-Net and trainable copies to enhance text-image alignment. Additionally, we introduce a $\textit{timestep-based weight scheduler}$ to relieve excessive control and increase the diversity of generation.
\item Extensive experiments show that our method can generate high-quality subject preserving images. Especially when there are multiple similar targets in the reference image, our method can accurately select the target subject and achieve the best results among all methods.

 \end{itemize}

\section{Related Work}
\label{gen_inst}

\textbf{Text-to-image diffusion models.} \quad
In recent years, diffusion models \cite{denoising-diffusion-model, diffusion-beat-gan, stable-diffusion} have developed rapidly due to their excellent image generation capabilities. The emergence of latent diffusion models has made it possible to quickly generate high-quality images and has achieved significant commercial success. StableDiffusion \cite{stable-diffusion}, DALL-E2 \cite{DALL-E2}, Imagen \cite{Imagen} trained on large-scale image-text pair datasets, have become mainstream in text-to-image generation. DiTs \cite{DiTs} replaces the U-Net in the diffusion model with transformers, further improving the text-to-image generation performance and scalability. The latest diffusion models, such as StableDiffusion-XL, Playground v2 \cite{playground-v2} can generate images with resolutions of $1024\times1024$ or even higher, and can accurately generate text in the images.

\textbf{Controllable generation.}\quad
In addition to using text prompts for conditional image generation, current diffusion models can add many additional inputs (e.g., edges, depth maps and segmentation maps) to achieve controllable image generation. ControlNet \cite{control-net} utilize a trainable copy of the U-Net to integrate multimodal information. It freezes the original U-Net and only fine-tunes the trainable copy to apply control as a plugin. IP adapter \cite{ip-adapter} takes the reference images as feature inputs and achieves image-to-image generation by injecting cross-attention between the reference image and the denoising image into the source model. Other methods \cite{diffedit, noisecollage, dragdiffusion} use DDIM inversion \cite{DDIM} to generate noised images and realize image editing by manipulate the noised images.

\textbf{Subject-driven generation.}\quad
Subject-driven image generation can be divided into two frameworks based on the methods: those that require test-time fine-tuning of the reference images and those that do not. For methods that require test-time fine-tuning, Text Inversion \cite{text-inversion} uses text vectors as optimizable parameters to search for specific vectors in the text space to generate similar images. Dreambooth \cite{dreambooth} fine-tunes the entire U-Net layers on reference images and proposes prior preservation loss to solve the language drift problem. Custom diffusion \cite{custom} and SVDiff \cite{svdiff} only fine-tune the most important parts of the U-Net, greatly reducing the number of parameters that need to be optimized. Some fine-tuning-free methods \cite{ELITE, taming, ip-adapter, Encoder-based-domain, fastcomposer, blip, ssrencoder, instantbooth} add additional adapters, using image encoder to encode reference images to embeddings as the subject representation. BLIP-Diffusion \cite{blip} can realize efficient zero-shot generation. SSR-Encoder \cite{ssrencoder} utilizes CLIP hidden state features to encode subject representation and can select target subject with text prompts or masks. ELITE \cite{ELITE} proposes global and local mapping networks for fast and accurate customized text-to-image generation. Instantbooth \cite{instantbooth} uses learnable image encoder and encodes input images as textual token, learning rich visual feature representations by introducing a few adapter layers to the pre-trained model. 

However, the above methods cannot accurately select the subject concepts that need to be learned or require additional text descriptions or expensive pixel-level masks. Our method uses two points to label the positive and negative regions in the reference image, enabling selective subject generation with minimal annotation cost.

\begin{figure*}
    \centering
    \includegraphics[width=1\linewidth]{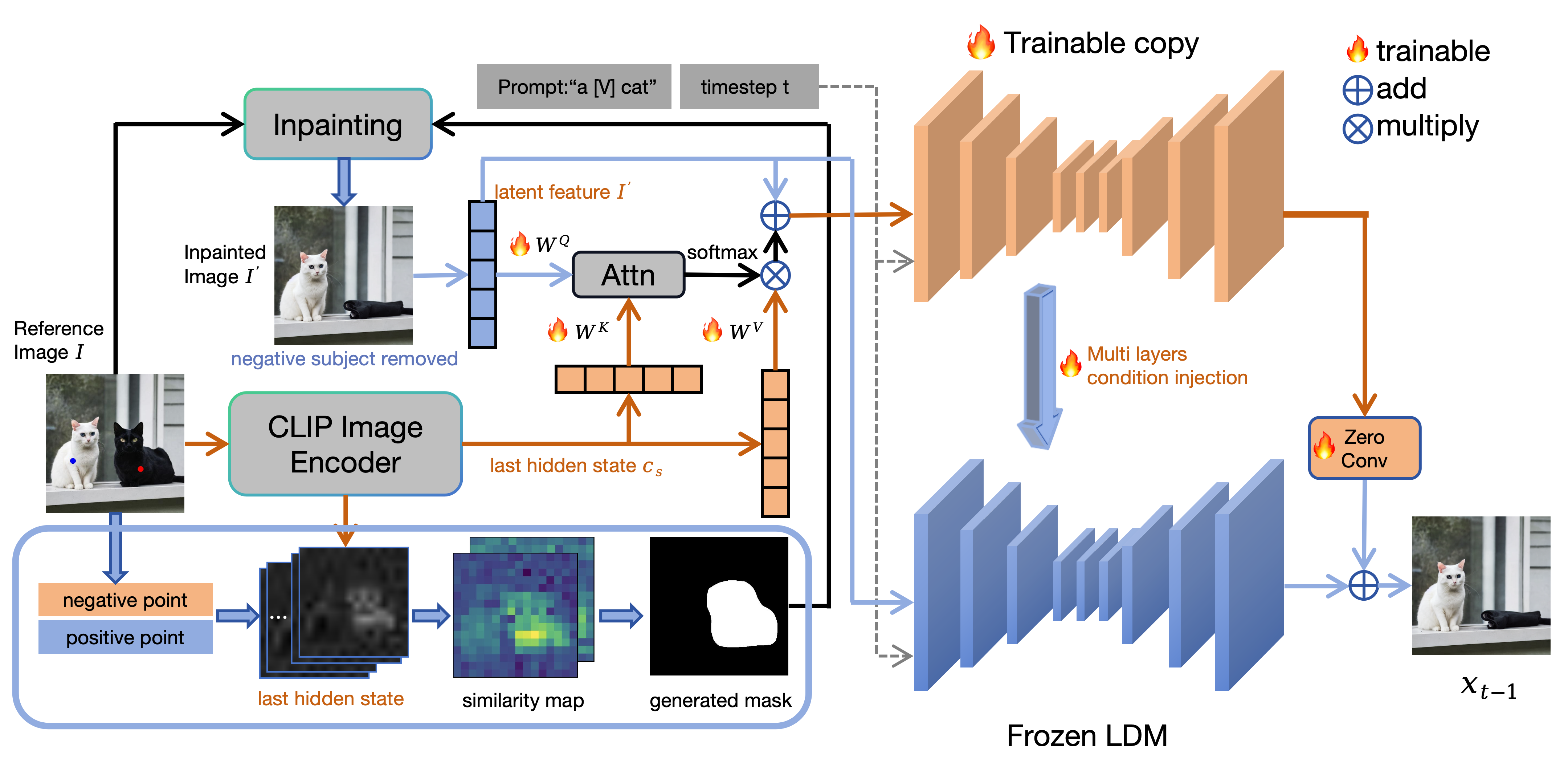}
    \caption{\textbf{Overview of our method}. Given point-image pairs $(p, I)$, the \OurEncoder will calculate the patch-level similarity on the hidden state feature from CLIP image encoder.It achieves obtaining a rough mask from point information without the need for any additional segmentation models.Then we inpaint the masked image and enhance subject presentation by image-image cross-attention.After that, the encoded feature serve as the input of the trainable copy and add conditions to the original U-Net by Multi-layers Condition Injection. }
    \label{fig:pipeline}
\end{figure*}

\section{The Method}
\label{method}

Subject-driven image generation aims to generate target images with high fidelity and creative editability according to given reference images. Usually, the content in an image is diverse, and relying solely on the provided text description make it difficult to accurately select the subject we want to retain. Compared to additional text prompts or expensive pixel-level masks, we propose \OurFramework, a specialized framework that can use a point to supervise model selection of corresponding features.

Formally, for a given reference image $I$ and a point $p$, the \OurEncoder effectively identifies the target subject and reduces negative information.The processed image $I$ is encoded to be a latent embedding and added to the original image embedding. The aligned feature is integrated into the original model with trainable copies of U-Net model layers. In the generation process, image embedding is added to the noise to give it a latent bias towards the subject. Enabling the model to have higher stability against different random noises and produce selected subjects with high fidelity. The overall methodology is illustrated in Fig. \ref{fig:pipeline}.

In general, our \OurFramework based on text-to-image diffusion models \cite{stable-diffusion}. It consists of two main parts: \OurEncoder (Sec. \ref{ssencoder}) and Multi-layers Condition Injection (Sec. \ref{subject-driven-generation}).

\subsection{Point Supervision Selective Subject Representation Encoder}
\label{ssencoder}
Considering that when there are multiple semantically similar subjects (like two different dogs) in an image, the diffusion model trained on this image may not be able to distinguish the features of different entities. This results in confusion in subject representation, causing the diffusion model to mix the different subjects. To solve this problem, we need to obtain a mask for negative information and remove irrelevant subject information from the image. However, it is difficult to segment point-annotated images without using additional pre-trained segment models. Several works \cite{custom, dreambooth} have proven that fine-tune a diffusion model with unique identifier will guide the model to focus on specific subjects without being influenced by the background. The minor inconsistencies and unrealistic parts in the training image will not affect the subject representation.

Thus, we only need to generate a rough mask to eliminate negative information and fill the mask to remove unrelated subjects. We utilize CLIP image encoder to encode the image to feature embedding and calculate the similarity between the patch where the point is located and all other patches. Mathematically, given a image $I$ point coordinates $p=\{x, y\}$, we can calculate the patch similarity map $M_{\{ Negative, Positive \}}\in \mathbb{R}^{N\times N}$ as follows:

\begin{equation}\label{eq0}
M_{\{N, P\}}=\{M_{i, j}, M_{i, j}=Cos(patch_{x^{'}, y^{'}}, patch_{i, j})\}.
\end{equation}

Considering that CLIP is trained on class-level annotation, it fails to accurately recognize objects with different instances of the same class. We utilize a negative similarity map to modify the positive similarity map and perform Gaussian filtering  to smooth the similarity map. We then use the Otsu method to binarize the similarity map and remove outliers that are not connected to the specified points:
\begin{equation}\label{eq1}
Mask= Otsu(\Psi(M_P*(1-M_N))),
\end{equation}

\begin{equation}\label{eq2}
I^{'} = inpainting(I, Mask),
\end{equation}

where $\Psi$ is a Gaussian filter and $Otsu$ is Otsu method.
After obtaining a rough mask of negative subject, we use an inpainting model \cite{stable-diffusion} to complete the masked image. In fact, the masked area can be filled with any content as long as it is not similar to the target subject. The feature extraction capability of the diffusion model allows it to focus on specific subject regions with the text prompts.

In order to move the generated image distribution in the latent space towards the reference image distribution and highlight the target subject, we encode the completed image into the latent space and compute cross-attention with the output of the last layer of the CLIP image encoder. Concretely, $I^{'}_{latent}$ is the latent image and $c_s$ is the last hidden state of CLIP image encoder. We create linear projections $W^Q$ for $I^{'}_{latent}$ as query and $W^K, W^V$ for $c_s$ as key and value.The total latent input can be formulated as Eq.\ref{latent_i}:

\begin{equation}\label{latent_i}
I^{''} = I^{'} + Softmax(\frac{Q_{i}K^{T}_{c}}{\sqrt{d}})V_{c},
\end{equation}

\begin{equation}\label{latent_i2}
 Q_i=W^Q\mathcal{Z}(I^{'}, \theta^z), K_c=W^Kc_s, V_c=W^Vc_s,
\end{equation}

where $I^{''}$ is the input of trainable copy and $\mathcal{Z}(·;\theta)$ is a $1\times1$ convolution layer with both weight and bias initialized to zeros.

\begin{figure}
    \centering
    \includegraphics[width=1\linewidth]{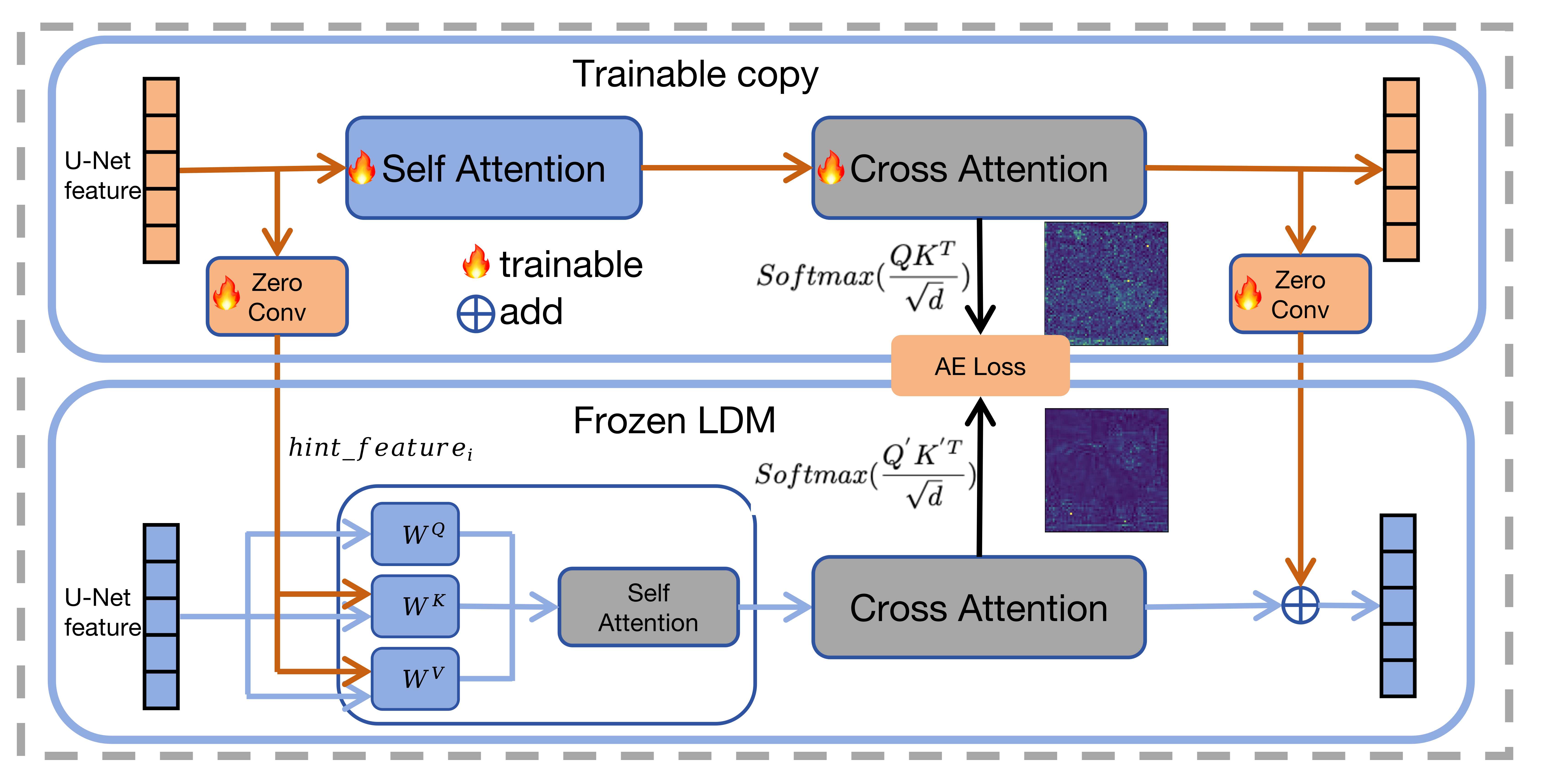}
    \caption{\textbf{Subject-driven Generation}. We utilize Multi-layers Condition Injection to add conditions. Specifically, we add the hidden states both in self and cross attention layers to the original model through a zero convolution. This can inject a detailed image representation and keep the generated subject. During training, we adopt denoising reconstruction loss $\mathcal{L}_{LDM}$ and attention consistency loss $\mathcal{L}_{ac}$ for consistent representation of attention of the selected subject.}
    \label{fig:injection}
\end{figure}

\subsection{Subject-driven Generation}
\label{subject-driven-generation}
In our approach, the encoded latent feature is projected into spatial-transformer blocks in the original U-Net for controllable generation. To achieve this, we copy the layers of a simplified U-Net (with only one ResNet block in each block), each corresponding to a spatial-transformer layer in the original U-Net. Inspired by \cite{custom} \cite{magicanimate}, self-attention and cross-attention layers have the greatest impact on the generation results of the model. The injection is composed of two parts: self attention condition injection and cross-attention enhancement, as shown on Fig. \ref{fig:injection}.

\textbf{Multi-layers Condition Injection.}\quad Our approach creates a trainable copy of the simplified U-Net $\mathcal{F}(x;\theta)$. For each denoising step $t$, we compute the hidden state features in each self-attention layer. Mathematically, the hidden state features can be represented in the following form:

\begin{equation}\label{eq4}
f_i=\mathcal{Z}(\mathcal{F}_i(x_t, c, t, \theta_i), \theta^{z}_{i}),
\end{equation}

where $f=\{f_1,f_2,...,f_n\}$ is a set of the self-attention hidden features and $n$ is the number of self attention layers. Different from Control-Net which add conditions to output blocks as residual connection, we deliver the hidden features to the self-attention layers in the original U-Net as follow:

\begin{equation}\label{eq5}
 Q=W^Qz_i, K'=W^K[z_i, f_i], V'=W^V[z_i+\lambda f_i, f_i],
\end{equation}

\begin{equation}\label{eq6}
 Attention = Softmax(\frac{QK^{'T}}{\sqrt{d}})V^{'},
\end{equation}

where $z_i$ is the hidden state feature in the original U-Net. We concatenate and add the additional features to the original self-attention layers to strengthen the presentation of the selected subject. The layers are controlled by a hyperparameter $\lambda$ = 0.2. This approach can not only preserves the detailed features in the reference images(such as dog's fur patterns, patterns on backpacks, etc.), but also make model pay more attention to the target subject during training, reducing conceptual confusion caused by the background. To accelerate convergence speed and reduce blurring in generated images, we also applied a zero convolution on the hidden features.

Prior research \cite{fastcomposer, pivotal, designing} shows that in the early stages of the denoising process, a rough structural layout of the image will be formed. Applying control to the diffusion model too early can result in the generated image matching the reference image very well, but causing low diversity. Different from the $\textit{delayed subject conditioning}$ \cite{fastcomposer} in prior works \cite{style-gan}, we propose a simple $\textit{timestep-based weight scheduler}$, which allows for arbitrary changes in control intensity during the inference phase, striking a balance between identity preservation and editability. Specifically, we use a hyperparameter $\varepsilon$ as the weight to perform conditional injection on the original U-Net, which uses only text prompts. Our $\textit{timestep-based weight scheduler}$ can be mathematically represented as:

\begin{equation}\label{eq7}
 \epsilon_t = \mathcal{F}(\varepsilon_t*f, x_t, c, t, \theta),
\end{equation}

\begin{equation}\label{eq8}
 \varepsilon_t = 1-\alpha(\frac{t}{T})^k+\beta,
\end{equation}

where $\epsilon_t$ is the predicted noise on the time-step $t$. $\alpha, \beta, k$ control multiplier, exponential descent, and bias separately. Empirically, we set $\alpha=0.5, \beta=0.2, k=2$ to balance prompt consistency and identity preservation.

In addition,we also explored other weight control methods in section \ref{ablation_study}.

\textbf{Attention Consistency Loss.}\quad
During the training phase, due to differences in input images, the cross-attention layers of the original model and the trainable copies may focus on different regions, leading to the model ignoring the detailed features of the reference image (such as color, pattern, etc.). To address this, we propose a novel Attention Consistency Loss $\mathcal{L}_{ac}$. This loss is designed to enhance similarity between the cross-attention map $M_{c}$ of the trainable copies and the one $M_{o}$
of the original U-Net. Specifically, we only choose the last cross-attention layer to calculate the MSE loss. $\mathcal{L}_{ac}$ can be mathematically represented as:

\begin{equation}\label{eq9}
 \mathcal{L}_{ac} = MSE(M_{c}, M_{o})/avg= Mean(||M_{c}, M_{o}||^2)/avg,
\end{equation}

\begin{equation}\label{eq10}
avg = (Sum(M_{c})+Sum(M_{o}))/2,
\end{equation}

\begin{equation}\label{eq11}
 M_{c} = Softmax(\frac{QK^{T}}{\sqrt{d}}), M_{o} = Softmax(\frac{Q^{'}K^{'T}}{\sqrt{d}}).
\end{equation}

Following the Stable diffusion model \cite{stable-diffusion}, we also utilize a latent denoising loss $\mathcal{L}_{LDM}$ to Minimize the KL divergence between the generated image distribution and the target image distribution, as demonstrated in Eq.\ref{eq12}:

\begin{equation}\label{eq12}
 \mathcal{L}_{LDM} = \mathbb{E}_{x_0, t, \epsilon}[||\epsilon - \epsilon_\theta(x_t, t, c_t)||^{2}_{2}],
\end{equation}

where $x_t$ is the noisy image at time step $t$, $\epsilon$ is the added noise and $\epsilon_\theta$ is the predicted noise in the parameters $\theta$.

Thus, our total loss can be presented as:

\begin{equation}\label{eq13}
 \mathcal{L} = \mathcal{L_{LDM}}+\gamma\mathcal{L}_{ac},
\end{equation}

where $\gamma$ is set as a constant, empirically set to 0.1. Meanwhile, since the original model is frozen, the changes in its cross-attention map will be smaller than those of the trainable copies. Therefore, minimizing their cross-attention map consistency loss can also make the model focus more on the target subject, thereby alleviating semantic drift.

\begin{figure*}
    \centering
    \includegraphics[width=1\linewidth]{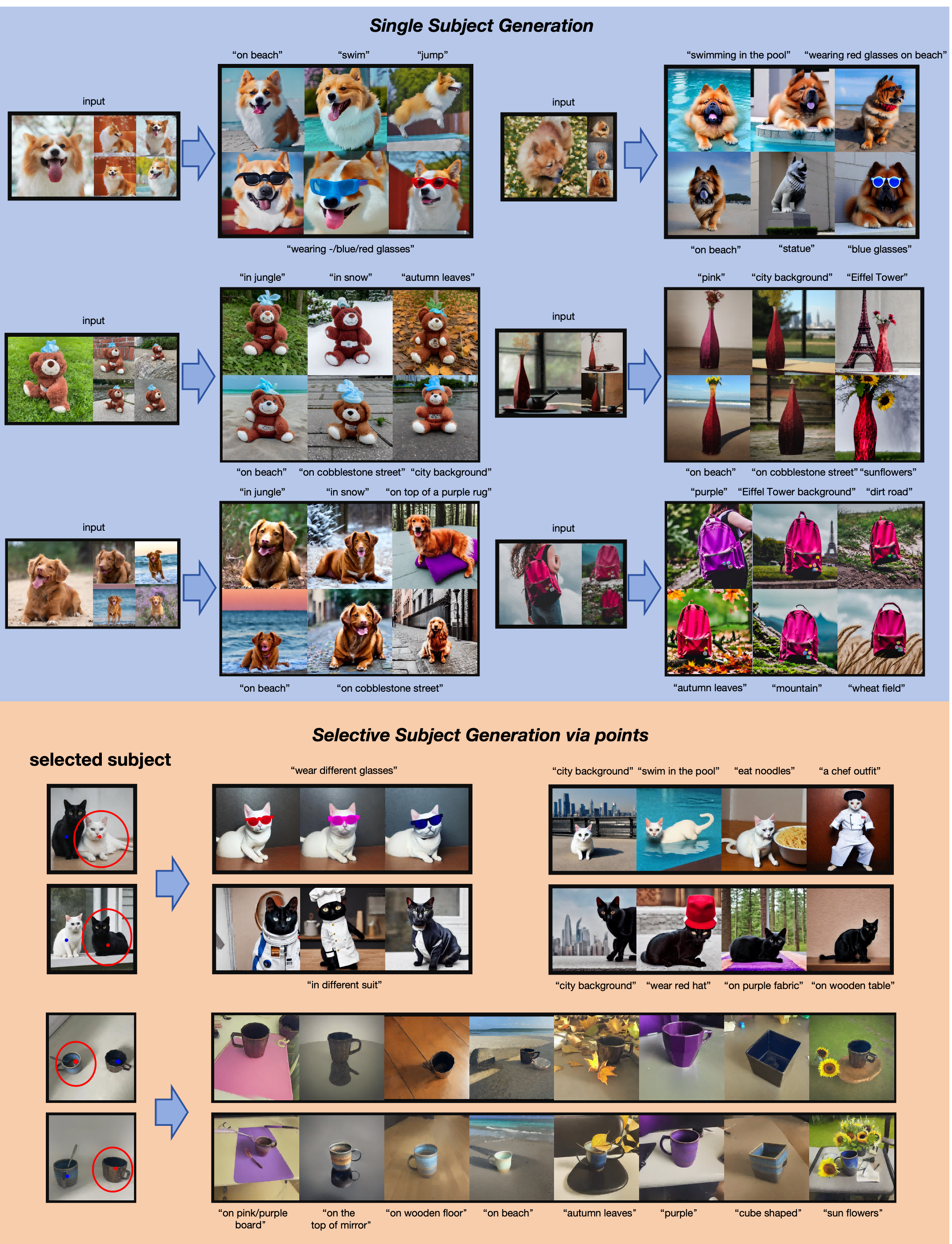}
    \caption{\textbf{Qualitative results of \OurFramework}.Our method is adaptable for both single subject generation and selective subject generation with points.}
    \label{fig:result}
\end{figure*}

\section{Experiments}
In this section, we present our experiments and training details. Our method can select the target subject in the reference image through two points and exclude negative subjects. Inspired by prior works \cite{dreambooth}, we also use a unique identifier $[V]$ to represent the target subject.
\subsection{Setup}
\textbf{Training Details.} \quad Our method is based on StableDiffusion v1-5 \cite{stable-diffusion} model. To encode the visual inputs and calculate the patch similarity, we use OpenAI's clip-vit-large-patch14 vision model, which can also encode text prompt in SD v1-5. For every input classes, we train our models for 60 epochs on a single NVIDIA RTX3090 GPU, with a learning rate of 1e-5. The training process takes about 6 minutes. During inference, we use DDIM as the sampler, with a step size of 50 and a guidance scale set to 7.5. To realize Classifier-Free Guidance, we drop out 10\% of the conditions.

\textbf{Evaluation Metrics.} \quad 
 To evaluate our method, we use DreamBench datasets \cite{dreambooth} which includes 30 categories, with 4-7 images per category. In order to evaluate the performance of the model in detail, we adopted the following evaluation indicators:DINO Scores \cite{DINO}, CLIP-I, CLIP-T, CLIP-Iv, DINO-v. CLIP-I is the cosine similarity between the CLIP embeddings of generated images and source images. Considering that CLIP focuses more on features of different categories and is not sensitive to differences between similar objects, we utilize GroundingDINOv2 to accurately determine image similarity. Image similarity is used to evaluate the fidelity of generated images. To evaluate the diversity, we use CLIP-T which calculate the cosine similarity between the CLIP embeddings of generated images and text prompts. CLIP-Iv and DINO-v represent the variance of the generated images' CLIP-I and DINOv2 scores, which can be used to measure the stability of generating images.

 \textbf{Comparisons.} \quad 
We compare our results with recent works using the hyperparameters provided in their work or the most commonly used codes. For each method, we set the same random seed before training and generating on each class. Following \cite{dreambooth}, we use 25 different text prompts and generate 4 images for each prompts. Each class will generate 100 images, with a total of 3000 images for 30 classes. We calculate the cosine similarity between each generated images and all reference images and take the average as the score for the image. 

\begin{table*}
  \caption{\textbf{Quantitative comparison}.Metrics that are bold and underlined represent methods that rank 1st and 2nd.}
  \label{comparison}
  \centering
  \begin{tabular}{llllll}
    \toprule
    Method     & CLIP-I$\uparrow$ & CLIP-T$\uparrow$ &DINOv2$\uparrow$ &CLIP-Iv $\downarrow$&DINO-v$\downarrow$\\
    \midrule
    DreamBooth \cite{dreambooth}      & 0.7511 & 0.2541& 0.4737 & 0.0078 & 0.0280\\
    CustomDiffusion \cite{custom} & 0.6832 & \bf{0.2762} & 0.3869 & 0.0066 & 0.0188\\
    BLIP-Diffusion \cite{blip}  & 0.7428 & 0.2207 & 0.5121 &0.0046 & \bf{0.0127}\\
    SSR-Encoder\cite{ssrencoder} & \bf{0.7793} & 0.2355 & 0.5323 & \bf{0.0032} & \underline{0.0140}\\
    \midrule
    Ours            & \underline{0.7748} & 0.2383 & \bf{0.5491} & \underline{0.0045} & 0.0187\\
    Ours W/ fixed weight&  0.7596 &	0.2420 &	\underline{0.5390} &	0.0049 &	0.0177\\
    Ours W/ $\mathcal{L}_{ac}$(all U-Net) &	0.6474 & \underline{0.2564} &	0.3762 & 	0.0050 & 	0.0144 \\
    
    Ours W/o $\mathcal{L}_{ac}$ &	0.7099 & 0.2506 &	0.4783 & 	0.0056 & 	0.0161 \\

    \bottomrule
  \end{tabular}
\end{table*}

\subsection{Experiment Results}
\textbf{Qualitative results.}\quad
Fig. \ref{fig:result} demonstrates the overall generated subject images. The result consists of two parts: single subject image generation and similar subject image generation. A large number of generated images have proven that our method can stably and effectively preserve the features of the target subject, and faithfully generate according to the given text prompts. Our method achieved stunning results in the single subject image generation. When generating subject images of similar subjects, relying on the supervision of positive and negative points, our method can accurately extract the features of the target subject without being affected by negative targets.

\textbf{Quantitative comparison.} \quad
 Table \ref{comparison} presents our quantitative comparison with other method on DreamBench. Overall, \OurFramework basically outweighs previous method, especially in subject alignment including CLIP-I and DINOv2. Our method has a great advantage in preserving the target subject features and can also ensure the diversity of generation. In terms of the stability of generated images including CLIP-Iv and DINO-v, our method achieves the highest CLIP-Iv and the second highest DINO-v scores. This shows that our method is less affected by random numbers and can stably generate subject images. Additionally, high image fidelity and low variance also prove that our method can stably and effectively generate high-fidelity subject images.

\textbf{Qualitative comparison.} \quad
For the problem of being unable to identify similar subjects, as shown on Fig.\ref{fig:point_comparison}, our method can accurately extract the features of the target subject using two points and faithfully generate it according to the reference image and the given text. Other methods cannot distinguish different subjects in the reference image and will incorrectly generate images with mixed features or multiple subjects.

In summary, our method can utilize point supervision to ensure an accurate representation of the selected image subjects. Thereby addressing key challenges faced by other current methods.

\begin{table}
  \caption{\textbf{Ablation results}. The impact of each part on the prompt consistency and identity preservation.}
  \label{ablation}
  \centering
  \begin{tabular}{llll}
    \toprule
    Method     & CLIP-I$\uparrow$     & CLIP-T$\uparrow$ &DINOv2$\uparrow$\\
    \midrule
    W/o \OurEncoder& 0.7117 & \bf{0.2602} & 0.4632\\
    W/o $\mathcal{L}_{ac}$ & 0.7099 & \underline{0.2506} & 0.4783\\
    W/  weight scheduler  & \underline{0.7596} & 0.2420 & \underline{0.5390} \\
    \midrule
    Ours(Full)            & \bf{0.7748} & 0.2383 & \bf{0.5491} \\
    \bottomrule
  \end{tabular}
\end{table}

\subsection{Ablation study.}
\label{ablation_study}
Our ablation study explore the impact of different modules on the fidelity and prompt consistency of generated images, as show on Table \ref{ablation}. We found that the \OurEncoder and $\mathcal{L}_{ac}$ make the generated image more in line with the subject representation, but reduce editability. The timestep-based weight scheduler helps improve prompt consistency but reduce identity preservation. We have analyzed the reasons for these phenomena below.

\begin{figure}
    \centering
    \includegraphics[width=1\linewidth]{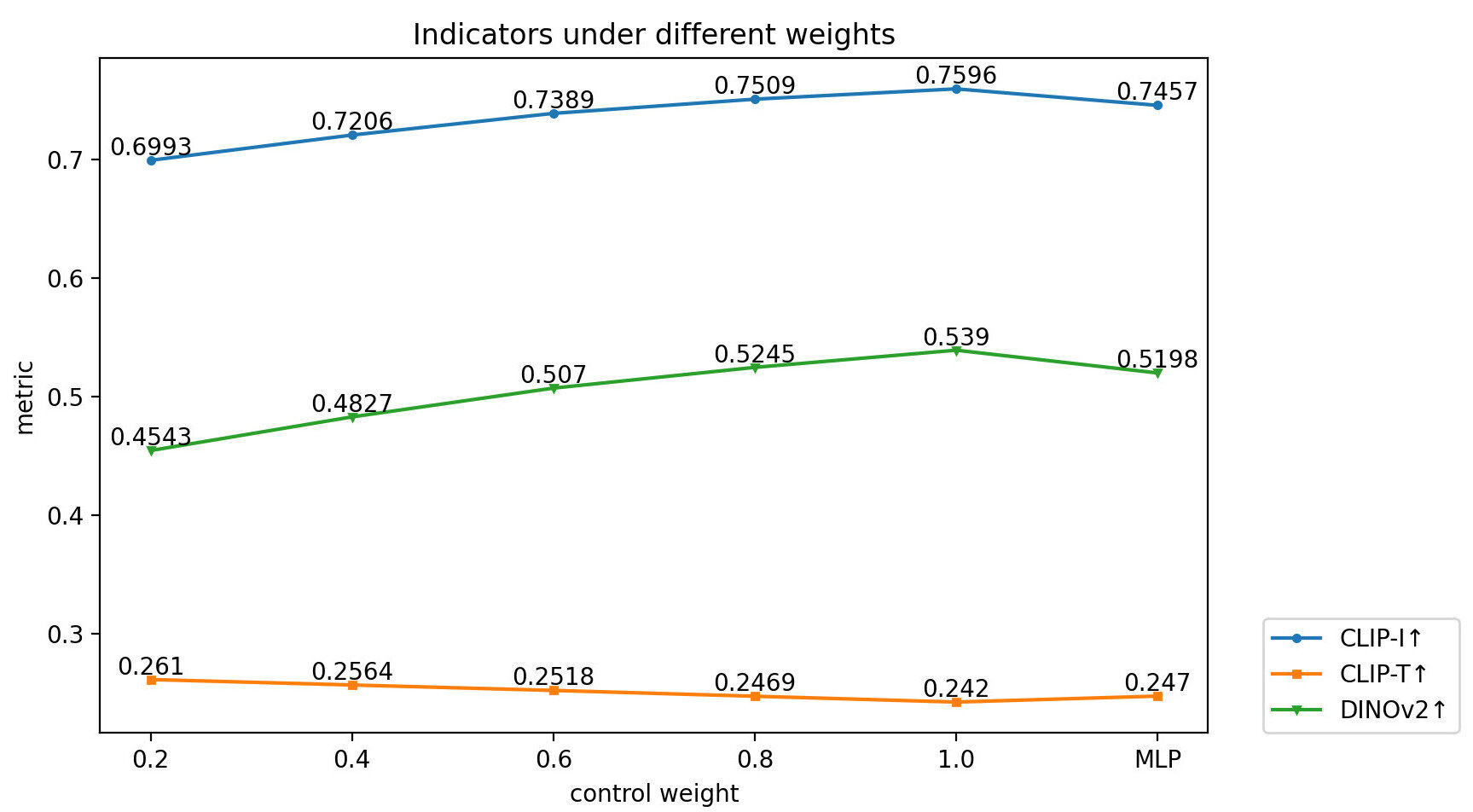}
    \caption{\textbf{Result of different weight}. Adjust the control weights between original U-Net and trainable copy will balance the prompt consistency and identity preservation. MLP represents a learnable control weight parameter.}
    \label{fig:different_weight}
\end{figure}

\begin{table}
  \caption{\textbf{A tiny experiment}. We naively use variable weights that increase or decrease quadratically with time steps.}
  \label{variable_weight}
  \centering
  \begin{tabular}{llll}
    \toprule
    Weight curve     & CLIP-I$\uparrow$     & CLIP-T$\uparrow$ &DINOv2$\uparrow$\\
    \midrule
    Increasing weight& 0.6991 & 0.2553 & 0.4531\\
    Decreasing weight & 0.7025 & 0.2393 & 0.4530\\
    \bottomrule
  \end{tabular}
\end{table}

\begin{table}
  \caption{\textbf{Different attention consistency loss}. We compare the impact of using $\mathcal{L}_{ac}$ in different layers.}
  \label{different_ac-loss}
  \centering
  \begin{tabular}{llll}
    \toprule
    $\mathcal{L}_{ac}$     & CLIP-I$\uparrow$     & CLIP-T$\uparrow$ &DINOv2$\uparrow$\\
    \midrule
    W/o $\mathcal{L}_{ac}$ &0.7099&0.2506&0.4783\\
    W/ the last layer& \bf{0.7596} & 0.2420 & \bf{0.5390}\\
    W/ the whole U-Net & 0.6474 & \bf{0.2564} & 0.3762\\
    \bottomrule
  \end{tabular}
\end{table}

\textbf{\OurEncoderTitle.} \quad 
We conducted ablation study on \OurEncoder, $\mathcal{L}_{ac}$, and \textit{timestep-based weight scheduler}. \OurEncoder can encode the hidden features of CLIP image encoder into the latent space representation of the image, and add a bias to the target image distribution, making its distribution distinct from other image distributions. This allows the model to generate more realistic images, and obtain a higher subject alignment score.

\textbf{Attention consistency loss.} \quad 
Attention consistency loss can align the attention of the trainable copy with the source model, allowing they to focus on the same area in the generated images. Additionally, $\mathcal{L}_{ac}$ makes the attention of the trainable copy close to the attention of the frozen and slower-changing source model, alleviating the semantic drift. We only use the attention map of the last layer in the U-Net to calculate $\mathcal{L}_{ac}$. Table \ref{different_ac-loss} shows the results of using $\mathcal{L}_{ac}$ in different layers.

This ablation study demonstrate that $\mathcal{L}_{ac}$ is able to significantly improve image fidelity while sacrificing a small portion of generated diversity.Using it on the last layer of U-Net yields the best results.While $\mathcal{L}_{ac}$ is applied to the whole U-Net,the increase in gradient will lead to slower convergence of the model and result in underfitting and difficulty in personalization generation.

\textbf{Timestep-based weight scheduler.} \quad 
Adjust the control weights between original U-Net and trainable copy will balance the prompt consistency and identity preservation as shown on Fig. \ref{fig:different_weight}. Besides using fixed weights, we also demonstrate the results using different weights at different time steps on Table \ref{variable_weight}. The quadratic decreasing weight uses a smaller weight in the early stage of denoising, which ensures the diversity of the generated image structure and can obtain a higher diversity score. The increasing weight is the opposite.

As shown on Fig. \ref{fig:different_weight},we find that as the control weight increases,the fidelity of the image continues to improve,but the editability decreases.We also try to utilize a trainable MLP to learn the optimal weight.However,it can only balance 
editability and fidelity.

Compared to applying control to U-Net at all time steps $t$, our schedule can effectively control the weights of injected conditions under different time steps. By giving lower weights in the early denoising stage and higher weights in the middle and later stages, our method can effectively balance prompt consistency and identity preservation. Although the image fidelity is very high without using timestep-based weight scheduler, it can cause the background of the generated image to be very similar to the reference image, as shown on \cite{fastcomposer}.

\section{Conclution}
 In this paper, we introduced the \OurFramework, a method of selective subject-driven generation via point supervision. In this method, we implemented a feature that was not present in previous methods: using only points to annotate the target subject allows the model to perform selective subject generation. Our method can effectively preserve the main features of the reference images and faithfully generate the images according to the given text prompts.

 However, our method also has some limitations. It is difficult to generate high-fidelity images of subjects with rich details. Additionally, our point supervision method does not perform well on non-salient images.

\bibliographystyle{unsrt}
\bibliography{reference}


\end{document}